\documentclass[conference]{IEEEtran}
\usepackage[english]{babel}
\IEEEoverridecommandlockouts
\usepackage{cite}
\usepackage{amsmath,amssymb,amsfonts}
\usepackage{algorithmic}
\usepackage{graphicx}
\usepackage{textcomp}
\usepackage{xcolor}
\begin{document}

\title{Minimum Class Confusion based Transfer for Land Cover Segmentation in Rural and Urban Regions \\ 
}

\author{\IEEEauthorblockN{Metehan Yalçın, Ahmet Alp Kındıroğlu, Furkan Burak Bağcı, Ufuk Uyan, Mahiye Uluyağmur Öztürk}

\IEEEauthorblockA{\textit\textit{Huawei Turkey R\&D Center} \\
Istanbul, Turkey \\
\{metehan.yalcin, ahmet.alp.kindiroglu, furkan.burak.bagci, ufuk.uyan1, mahiye.uluyagmur.ozturk\}@huawei.com
}}

\maketitle

\begin{abstract}

Transfer Learning methods are widely used in satellite image segmentation problems and improve performance upon classical supervised learning methods. In this study, we present a semantic segmentation method that allows us to make land cover maps by using transfer learning methods. We compare models trained in low-resolution images with insufficient data for the targeted region or zoom level. In order to boost performance on target data we experiment with models trained with unsupervised, semi-supervised and supervised transfer learning approaches, including satellite images from public datasets and other unlabeled sources.
According to experimental results, transfer learning improves segmentation performance 3.4\% MIoU (Mean Intersection over Union) in rural regions and 12.9\% MIoU in urban regions. We observed that transfer learning is more effective when two datasets share a comparable zoom level and are labeled with identical rules; otherwise, semi-supervised learning is more effective by using the data as unlabeled. In addition, experiments showed that HRNet outperformed building segmentation approaches in multi-class segmentation.

\end{abstract}


\section{Introduction}
Maps of land cover provide spatial data on many types of physical surface cover, such as a village, forests, croplands, lakes, and roads. Thanks to information obtained from land cover maps, many necessary applications such as city and regional planning, agricultural planning, communication and transportation infrastructure planning can be easily made. For the detailed extraction of these maps, methods such as recording with vehicles with three-dimensional scanners and sensors, combining images from manned and unmanned aerial vehicles and using satellite images are actively used. The most detailed and high-quality data of these are considered ground-level scanning. However, it is very difficult to obtain data with this method, especially in many areas where vehicle access is not possible, such as rural areas, coasts, islands and  mines. For this reason, in recent years, efforts to obtain earth-use maps from satellite images at a cheap cost have gained speed. \\

Machine learning methods use the category of each pixel of images as label information to solve segmentation problems in satellite imagery. Thus, with the training set obtained, a pixel-based classification model can be trained, and prediction can be performed on the test data. However, in order for this process to be carried out successfully, there must be enough data with quality labels in the training set. In addition, it is important for training data and annotations to have content similar to the target problem data in order to achieve good performance.\cite{druck-etal-2009-active} \\

Since the data sets collected in remote sensing problems depend on the variety of satellite sensors and unmanned aerial sensors, standardization of annotations, and so on, it is quite challenging to find enough data that can be used for the desired problem. Utilizing different public datasets will reduce dataset dependency. Training the model over different examples with datasets from multiple sources is a very effective method in terms of achieving a higher capacity and general comprehensive model. \\

In this study, we aimed to produce low-cost land cover maps to be used in studies to improve the telecommunication infrastructure. The main difficulty in obtaining these maps is the difficulty of labeling and recognizing the resulting labels and images due to the low resolution of satellite images of these regions. For this purpose, we propose methods such as semi-supervised learning, transfer learning and combined learning to improve prediction success in low-resolution target data using city images and detailed images from other databases. In this method, we developed a model that extracts a land cover map using 20-meter resolution satellite images. We performed detailed hyperparameter analysis on this model with different architectures. Then we trained the model for 20-meter resolution dataset using 1 and 2-meter resolution datasets with semi-supervised learning and transfer learning approaches. We presented an analysis of these methods in our work.

\section{Related Works}
Segmentation is an image processing problem in which objects in the input image are estimated at the pixel level. Semantic segmentation aims to assign the same label to all objects belonging to the same class. In particular, studies in this field have gained momentum with the emergence of deep neural networks and annotated large datasets.  Fully Convolutional neural networks \cite{long2015fully} (FCN) was the first method to effectively apply deep neural networks to the semantic segmentation problem. In the FCN method, the outputs obtained from the neural network layers of different sizes were collected with the result obtained from the lowest layer and quality image outputs of the object edges were obtained. \\

U-Net \cite{ronneberger2015u}, which has a similar structure, is one of the most well-known models in image segmentation. The U-Net consists of two parts, the encoder and the decoder. While the encoder part extracts spatial-invariant features in the traditional CNN structure, The decoder part samples these features to the output image that is the same size as the input image. The difference between U-Net from FCN is that it uses a multi-channel structure in the decoder section and delivers the semantic information to high-resolution layers. Unet++ \cite{zhou2018unet++} method uses skip connections from shallow layers to deep layers that combine low and high-level features for better predictions.\\

Deeplab method \cite{chen2014semantic}, with its atrous convolutional layer, extended the receptive field of the models, allowing them to train better quality segmentation models with fewer parameters. In the following years, many improvements were made to this model; DeeplabV2 \cite{chen2017deeplab}, which can segment objects at different scales more consistently with ASPP (Atrous Spatial Pyramid Pooling), DeepLabV3 \cite{chen2017rethinking} with improvements on AC (Atrous Convolution) units, DeeplabV3+ \cite{chen2018encoder} which developed an efficient decoder module to improve segmentation results along object boundaries on to DeeplabV3 model  were developed. \\

Identifying specific objects in satellite images by semantic segmentation is an actively studied research area \cite{neupane2021deep}. The complex nature of the backgrounds in satellite images and the high in-class variability of objects (building, city, road etc.) are among the main challenges of the segmentation problem. However, there has been much progress in this regard with the ability to obtain high-resolution satellite images and the development of deep learning-based semantic segmentation methods. \\

Many U-Net-based models \cite{guo2018semantic, li2019semantic} have been proposed for the urban image segmentation problem. The DeepResUnet \cite{yi2019semantic} proposed by Yi et al. for building segmentation in urban areas has fewer parameters and performs better than the original U-Net, but the inference time is longer. Combining DeepLabV3+ \cite{chen2018encoder} with object-based image analysis, Du et al. \cite{du2021incorporating} achieved successful results in very high-resolution satellite images. In that study, classification results obtained using a DeepLabv3+ network on the spectral image and a random forest classifier on the features obtained by image analysis were converted to final estimates with conditional random field. In many recent studies, transformer-based models  \cite{hong2021spectralformer, xu2021efficient, scheibenreif2022self} have been successfully applied to the semantic segmentation problem. \\

Recently Wang et al. \cite{hrnet} created HRNet outperforming semantic segmentation architecture. They maintained classical encoder-decoder architecture to increase high resolution representations. The classical encoder-decoder approach first encodes the input image as low-resolutional representation, then the decoder part process this low resolution representation to high-resolution representation. HRNet contributes information exchange between these two parts parallelly, not in series as in the classical approach. \\

Geonrw \cite{baier2020building}, Inria\cite{maggiori2017dataset} and Deepglobe\cite{demir2018deepglobe} can be given an example of remote sensing datasets created to make land cover maps. Geonrw is a data set of RGB and SAR remote sensing images mostly taken from urban areas by synthesizing with a generative adversarial network. In addition to having 10 image segmentation classes, Geonrw is suitable for use in altitude estimation or semantic image synthesis applications from aerial photographs. \\
Transfer learning refers to methods developed to use labeled data or model for one or more classes on a different but related task. The studies carried out in this field are summarized in the literature review by Wang et al.\cite{wang2018deep}. If the feature space and the intended task are the same, but there are differences between the feature distributions, supervised learning transfer is used \cite{Zhang2019}. If the target and source features are different, or if the data annotations are suitable for use on different tasks, then the semi-supervised \cite{ouali2020overview} or unsupervised transfer learning \cite{wilson2020survey} methods are used for use on the target task.\\

Studies on supervised learning transfer focus on field adaptation methods such as feature space alignment, adding loss functions that provide space adaptation to the model error function, or adding capacity to the model architecture to store shared and discrete model parameters between target and source tasks separately. With the popularization of deep neural networks, the most popular method of transfer learning, which is widely used, is a fine-tuning method that uses weights of a trained model at initial weights of a new model  \cite{hinton2007backpropagation}. Studies on supervised transfer learning are divided by three into diversity-based, contrast-based and reconstruction-based by Zhang \cite{Zhang2019}. Examples of the most popular difference-based methods are those that transfer between tasks using class information \cite{jin2020minimum}, methods that measure statistical distribution difference with distance calculations \cite{long2017deep} such as MMD (maximum mean discrepancy), and various transferable methods such as domain-specific layers and normalizations \cite{li2016revisiting} in the model architecture. \\

Semi-supervised transfer learning is used successfully where large amounts of unlabeled or labeled data are available with a very different distribution \cite{ouali2020overview}. The methods used in this field can be examined in three groups as methods that perform continuity regulation, methods that produce a common pseudo task or pseudo label based methods and generative methods. In continuity regulation based methods, two different basic methods can be examined. The Minimum Class Confusion method \cite{jin2020minimum} is a method that increases the stability of the model over the unlabeled data by changing the model weights to make more stable predictions if the prediction of the model over the unlabeled data is not stable. Another approach uses methods such as randaugment  \cite{cubuk2020randaugment}, which make small and insignificant changes to the labels. Minimizing the impact of these changes on class predictions is a popular method among semi-supervised transfer learning methods. Unlike these methods, an example of false label generation-based methods can be given as the noisy student method \cite{xie2020self}.

\section{Method}
\subsection{Semantic Segmentation}\label{AA}
\begin{figure}[htbp]
\centerline{\includegraphics[width=1.0\columnwidth]{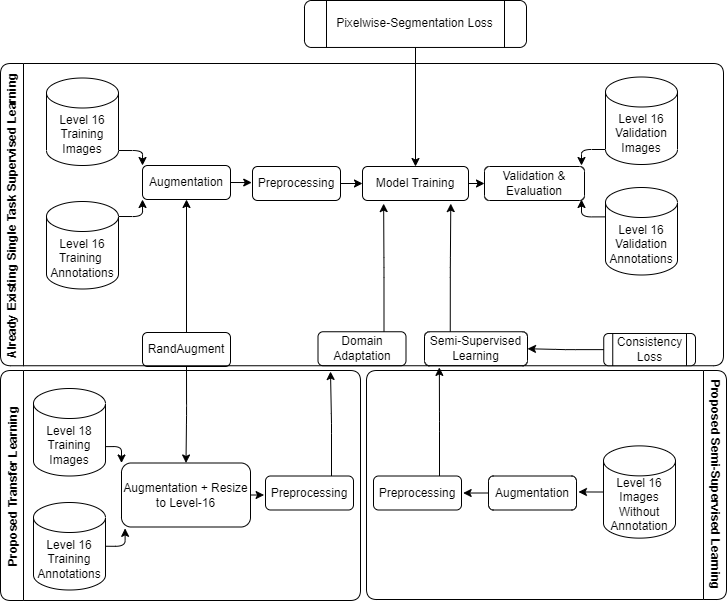}}

\caption{Semantic Segmentation Scheme}
\label{fig:sema}
\end{figure}

The extraction of land cover maps from satellite images is defined as a semantic segmentation problem. In this study, four different semantic segmentation architectures HRNet, Unet++, DeeplabV3+ and BisenetV1 architectures were used together with ResNet and Efficient-Net encoders.HRNet uses its own encoder backbone. The flowchart of our semantic segmentation method is shown in Figure \ref{fig:sema}. \\

A large part of the satellite images given as input to the spatial segmentation methods that we use consists of 5000x6000x3 channel RGB color pictures and labels of these pictures in the same size. In the first step of our training and testing method, a hyper-parameter optimization study was performed on these images for the training, and as a result, all experiments were performed on 512x512 randomly cropped images taken over 1000x1000 images extracted in grid form from large images. In addition test is performed 512x512 sliding window method on 5000x6000 test images. \\
Data augmentation is applied to cropped images to increase data diversity and to make the trained models more robust to effects such as light, color, displacement and rotation. The randaugment \cite{cubuk2020randaugment} method is used for data augmentation. In this method, increment methods such as translation, resizing, affine rotation, affine shearing, reversing the image, and random horizontal or vertical rotation It was chosen according to our problem. During the training, two of the data augmentation methods are randomly selected and applied to the image. In this method, augmentation methods such as translation, resizing, affine rotation, affine shearing, reversing the image, and random horizontal or vertical rotation is chosen according to our problem. During the training, two of the data augmentation methods are randomly selected and applied to the image. The degree of rotation is determined by taking 10 equal sections between 0 and 150. As the training step progresses, the rotation limit is increased and random selection is made from this section again. Following the data augmentation step, the images in the dataset are normalized with the mean and standard deviation values of the imagenet dataset. \\

The Unet++ architecture, which is the first of the methods used during the training, consists of encoder and decoder structures similar to the letter U. Unet++ \cite{zhou2018unet++} added an array of nested convolution blocks prior to merging to the skip connection where the same size feature maps between encoder and decoder are transferred to reduce semantic information loss due to the separation of the encoder and decoder paths in the UNet structure, and updated these jumps to be more intense. The second architecture that we use, the DeeplabV3+ method, uses methods such as atrous convolution and feature pyramiding compared to the Unet++ method. For this reason, it has advantages, in particular, in the effect of objects to be recognized by a wider area of influence on the image or in recognition of objects of different sizes. \\

Resnet50, Efficientnet B3 and Efficientnet B5 architectures are used as encoders in our experiments. The Efficientnet architecture \cite{tan2019efficientnet} is divided into two parts, convolution, point and depth convolutions, thus reduces the computational cost. It was chosen especially because of the ease with which the model complexity can be studied parametrically. HRNet architecture changes classical encoder decoder architecture. This model uses parallel encoder decoder path that shares representations paralelly using multi-scale paralell convolutions. HRNet model achieves best building segmentation results in our experiments in urban cities.

\subsection{Transfer Learning with Minumum Class Confusion}\label{AA}
One of the main problems we encountered during the semantic segmentation training for satellite images is that the labels used in education are not of sufficient quantity and quality. In this study, the combined training method in which data from different sources are used at the same time in training in order to learn data from different sources together. Semi-supervised transfer teaching and learning methods using the Minimum Class Confusion (MCC) method \cite{jin2020minimum} is used. In the common training methodology used for all three methods, the data from the target and source datasets are sampled to each form half of the mini-batch during the deep neural network training. In the co-learning method, the total loss function is calculated over the signs $y_t$ and $y_s$ of the target data $x_t$ and the source data $x_s$, as seen in Equation \ref{eq:1}. In the formula, the class label $L_{SN}$ represents the loss function, and $f(x,w)$ represents the model output obtained with x data. And also $w$ is learnable model parameters.

\begin{equation}
\resizebox{.8\hsize}{!}{$L_\varepsilon\bigl(y,f(x,w)\bigr) = L_{SN}\bigl(y_t,f(x_t,w)\bigr) + L_{SN}\bigl(y_s,f(x_s,w)\bigr)$} 
\label{eq:1}\end{equation}

The Minimum Class Confusion technique is a method that aims to minimize the classification error to the source data by converging the binary class confusion values of the estimates. This method inputs the output of a mini-batch of logits from the model and multiplies it by its inverse to obtain a correlation matrix that converges to the class confusion matrix. The process of unknown re-weighting is applied to normalize probabilities and to reflect more weight of important samples. Following this, a category normalization is performed in mini-batches to minimize the impact of the number of classes available on weights. In the matrix obtained at the end of these operations, classes showing high confusion with each other give higher correlation results. Since the values on the diagonal of the resulting matrix will represent the confusion of each class with itself, it will enable the model, which uses the loss function that aims to maximize these values or minimize the remaining values, to make more stable predictions on unlabeled data.  \\

The MCC method in semi-supervised training, the MCC loss function $ L_{MCC}$ is used for the source data, and the class label loss function is used for the current target data, as seen in Equation \ref{eq:2}.

\begin{equation}
\resizebox{.8\hsize}{!}{$L_\varepsilon\bigl(y,f(x,w)\bigr) = L_{SN}\bigl(y_t,f(x_t,w)\bigr) + L_{MCC}\bigl(y_s,f(x_s, w)\bigr)$} 
\label{eq:2}\end{equation}

In the MCC based transfer learning, both the MCC loss function for the source data and the class label loss function for the whole data set are used as shown in Equation \ref{eq:3}.

\begin{equation}
\resizebox{.8\hsize}{!}{$L_\varepsilon\bigl(y,f(x,w)\bigr) = L_{SN}\bigl(y,f(x,w)\bigr) + L_{MCC}\bigl(y_s,f(x_s, w)\bigr)$} 
\label{eq:3}\end{equation} 

\section{Experiments and Results}
\subsection{Datasets}\label{AA}
\begin{figure}[htbp]
\centerline{\includegraphics[width=0.8\columnwidth]{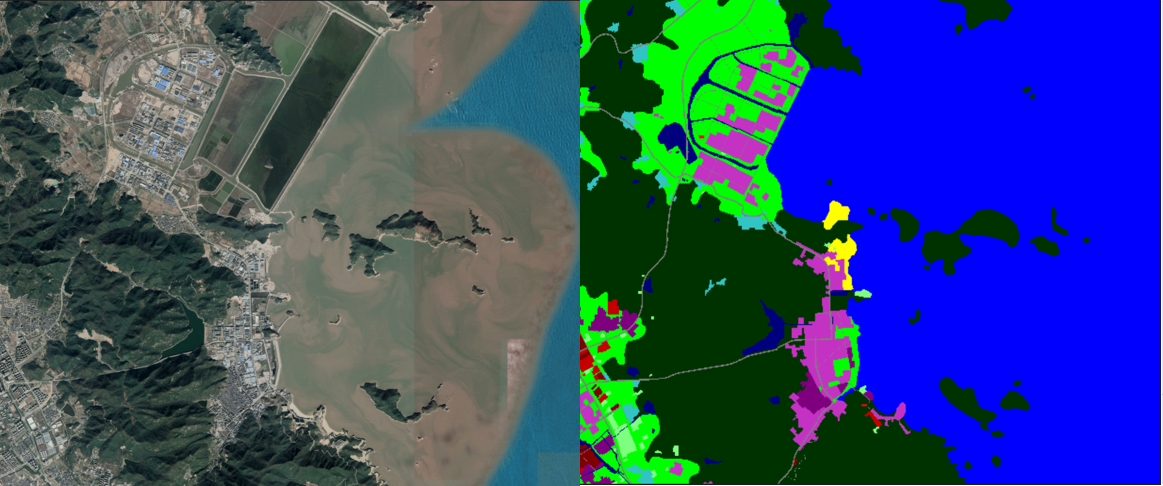}\includegraphics[width=0.2\columnwidth]{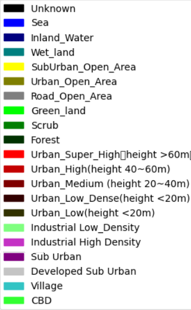}}
\caption{Example rgb image and 21-class semantic segmentation labels of dataset HWLC16}
\label{fig:legend}
\end{figure}

In this study, the Huawei Land Cover datasets HWLC16 and HWLC18 were created by Huawei, and the publicly available Geonrw and DeepGlobe datasets were used. The HWLC16 and HWLC18 datasets consist of satellite images taken from satellites by Huawei. The resulting images are at level 16 and level 18 according to OpenStreetmaps zoom levels \cite{Openstreetmaps:2022}. The HWLC16 dataset consists of images with a resolution of 2.5 meters per pixel obtained from rural areas. This dataset contains 315 5000*6000 training and 110 test images with the same resolution, and the labels have a lower resolution than the images. Each image has 21 classes of semantic segmentation labels with a resolution of 20 meters per pixel. Examples, class names and colorings of this dataset are given in Figure \ref{fig:legend}. This dataset was collected from rural areas where high-resolution satellite imagery is difficult and costly to obtain. The images used as test sets are divided into two as rural and urban regions, the set on the city region is called $urban\_test$ in the tables, and the set on the rural region is called $rural\_test$.

The second dataset, HWLC18 dataset, consists of level 18 satellite images, which is a more detailed zoom level compared to the HWLC16. Each image is composed of RGB color images with a resolution of 0.5 meters, while the labels have a quality of detail of 2 meters. In the study, 413 pieces of 5000*6000 pictures are utilized. The HWLC18 dataset consists of images of metropolitan cities and their surroundings, not rural areas, unlike the level 16 HWLC16 dataset. The labels have much more detailed and high quality, but due to the difficulty and cost of labeling at this level of detail, the total covered area is approximately 15 times smaller than the HWLC16 database. \\

Apart from these two datasets, Geonrw \cite{geonrw-20} and DeepGlobe \cite{demir2018deepglobe} datasets are also used in the study. The Geonrw database consists of 7783 images with a resolution of 1 meter per pixel, in 1000*1000 dimensions, and was created by combining images taken from aircraft of various cities in Germany and includes 10 different classes. The training set of DeepGlobe dataset contains 1000*1000 7227 images, and each pixel corresponds to 0.05 meters resolution. This dataset has 7 different classes.\\

All supervised learning methods used in this study are made by reducing them to 5 classes, which are the subclasses of the data used above. These classes are in order: $['unknown',\color{red}'urban',\color{yellow}'open\_area',\color{blue}'water',\color{green}'forest'\color{ black}]$. 


\subsection{Implementation Details, Evaluation Criteria and Experiments}\label{AA}

In this study, all the experiments are performed using Pytorch. The model structures and weights in the SMP library  \cite{Yakubovskiy:2019} are taken as initial values.  In the experiments, Nvidia 2080TI graphics cards are used for training with Automatic Mixed Precision. The model performance is calculated with the intersection over union (IoU) value calculated as  described in Equation \ref{eq:4} for all classes in each image in the experiments. The average IoU value for each image is found by averaging the IoU scores of all other classes except the unknown class of that image.
The total performance measurement is made on the dataset by taking the average these values for all test pictures.   

\begin{equation}
\resizebox{.4\hsize}{!}{$IoU = \frac{TP}{TP+FP+TN}$} \label{eq:4}
\end{equation} 

The first experiment was conducted to determine the most appropriate semantic segmentation architecture to be used for the extraction of land cover maps in this study. In this context, DeeplabV3+, Unet++ and BisenetV1 algorithms are tested with the Resnet50, EfficientnetB3 and EfficientnetB5 architectures and their model capacities and suitability for the problem are compared. The test results are shown in Table \ref{tab:tab1}.

\begin{table}[htbp]
\caption{The results of the experiments performed on the HWLC16 dataset with the architectures of different models.}
\begin{center}
\resizebox{.99\hsize}{!}{$
\begin{tabular}{|c|c|ccccc|c|}
\hline
\textbf{Model} & \textbf{Encoder} & \multicolumn{5}{c|}{\textbf{Rural Area Dataset}}                                                                                                                 & \textbf{City} \\ \hline
               &                    & \multicolumn{1}{c|}{\textbf{mIoU}}  & \multicolumn{1}{c|}{\textbf{Urban}} & \multicolumn{1}{c|}{\textbf{Open Area}} & \multicolumn{1}{c|}{\textbf{Forest}} & \textbf{Water}    & \textbf{Building}             \\ \hline
BisenetV1      & Resnet18           & \multicolumn{1}{c|}{0.558}          & \multicolumn{1}{c|}{0.55}           & \multicolumn{1}{c|}{0.372}            & \multicolumn{1}{c|}{0.807}          & 0.504          & 0.721                     \\ \hline
Unet++          & Resnet50           & \multicolumn{1}{c|}{0.5874}         & \multicolumn{1}{c|}{0.6128}         & \multicolumn{1}{c|}{0.3636}           & \multicolumn{1}{c|}{0.7564}         & 0.6167         & 0.76                      \\ \hline
Unet++          & EfficientnetB3    & \multicolumn{1}{c|}{\textbf{0.602}} & \multicolumn{1}{c|}{0.595}          & \multicolumn{1}{c|}{0.339}            & \multicolumn{1}{c|}{0.84}           & \textbf{0.634} & \textbf{0.796}            \\ \hline
Unet++          & EfficientnetB5    & \multicolumn{1}{c|}{0.592}          & \multicolumn{1}{c|}{0.582}          & \multicolumn{1}{c|}{0.324}            & \multicolumn{1}{c|}{0.846}          & 0.618          & 0.787                     \\ \hline
Deeplab+       & Resnet50           & \multicolumn{1}{c|}{0.458}          & \multicolumn{1}{c|}{0.471}          & \multicolumn{1}{c|}{0.249}            & \multicolumn{1}{c|}{0.704}          & 0.331          & 0.703                     \\ \hline
Deeplab+       & EfficientnetB3    & \multicolumn{1}{c|}{0.5856}         & \multicolumn{1}{c|}{0.584}          & \multicolumn{1}{c|}{0.3375}           & \multicolumn{1}{c|}{0.8267}         & 0.594          & 0.68                      \\ \hline
Deeplab+       & EfficientnetB5    & \multicolumn{1}{c|}{0.592}          & \multicolumn{1}{c|}{0.582}          & \multicolumn{1}{c|}{0.323}            & \multicolumn{1}{c|}{0.846}          & 0.617          & 0.762                     \\ \hline
\end{tabular}\label{tab:tab1}$}
\end{center}
\end{table}

It is experienced that the model trained using the Unet++ architecture and the efficientnet encoder showed the most successful results in both the HWLC16 rural area test set and urban area dataset. The capacity of this model is higher than the BiseNetv1 model and smaller than the DeepLabV3+ model. When the outputs of different models are examined, it seems that the U-Net model is able to find sharper boundaries in segmenting small objects on the image, especially buildings and inner-city neighborhoods. The failure to train more complex models such as Deeplab+ and EfficientNetB5 Encoder with the same success may be due to lack of data, the excess of annotation errors in the dataset, or the fact that the training process is not long enough to find the best parameters. \\

Following the architectural analysis, the capacity and best performance of the model were measured in training using a single database. To improve this performance, combined training, semi-supervised learning and transfer learning methods were applied to use information from other datasets with similar data, and their improvements were compared. The results of these experiments are given in Table \ref{tab:tab2}. \\

\begin{table*}[htpb]
\caption{The table gives results on two test sets of the target HWLC16 dataset using different source data. The HWLC18-$>$16 dataset contains images 4 times smaller than HWLC18.}
\begin{center}
\resizebox{0.95\textwidth}{!}{
\begin{tabular}{|c|c|ccccc|c|}
\hline
\textbf{Source Dataset}    & \textbf{Transfer Method} & \multicolumn{5}{c|}{\textbf{Rural Region Dataset}}                                                                                                               & \textbf{Urban Area Dataset} \\ \hline
                        &                           & \multicolumn{1}{c|}{mIoU}           & \multicolumn{1}{c|}{Urban}          & \multicolumn{1}{c|}{Open Area}        & \multicolumn{1}{c|}{Forest}          & Water             & Building                      \\ \hline
-                       &                           & \multicolumn{1}{c|}{0.602}          & \multicolumn{1}{c|}{0.595}          & \multicolumn{1}{c|}{0.339}          & \multicolumn{1}{c|}{0.84}           & 0.634          & 0.796                     \\ \hline
HWLC18                  & Combined Training                  & \multicolumn{1}{c|}{0.617}          & \multicolumn{1}{c|}{0.567}          & \multicolumn{1}{c|}{0.391}          & \multicolumn{1}{c|}{0.87}           & 0.641          & 0.812                     \\ \hline
HWLC18                  & MCC Semi Supervised           & \multicolumn{1}{c|}{0.632}          & \multicolumn{1}{c|}{0.614}          & \multicolumn{1}{c|}{0.405}          & \multicolumn{1}{c|}{0.849}          & 0.66           & 0.869                     \\ \hline
HWLC18                  & MCC Transfer Learning                     & \multicolumn{1}{c|}{0.593}          & \multicolumn{1}{c|}{0.545}          & \multicolumn{1}{c|}{0.371}          & \multicolumn{1}{c|}{0.852}          & 0.606          & 0.518                     \\ \hline
HWLC18-\textgreater{}16 & Combined Training                  & \multicolumn{1}{c|}{0.616}          & \multicolumn{1}{c|}{0.62}           & \multicolumn{1}{c|}{0.4}            & \multicolumn{1}{c|}{0.857}          & 0.586          & 0.841                     \\ \hline
HWLC18-\textgreater{}16 & MCC Semi-Supervised           & \multicolumn{1}{c|}{0.615}          & \multicolumn{1}{c|}{0.621}          & \multicolumn{1}{c|}{0.395}          & \multicolumn{1}{c|}{0.844}          & 0.6            & 0.861                     \\ \hline
HWLC18-\textgreater{}16 & MCC Transfer Learning                    & \multicolumn{1}{c|}{\textbf{0.636}} & \multicolumn{1}{c|}{\textbf{0.621}} & \multicolumn{1}{c|}{\textbf{0.422}} & \multicolumn{1}{c|}{\textbf{0.861}} & 0.64           & 0.901                     \\ \hline
Geonrw                  & Combined Training                  & \multicolumn{1}{c|}{0.62}           & \multicolumn{1}{c|}{0.616}          & \multicolumn{1}{c|}{0.372}          & \multicolumn{1}{c|}{0.826}          & 0.666          & 0.842                     \\ \hline
Geonrw                  & MCC Semi-Supervised          & \multicolumn{1}{c|}{0.573}          & \multicolumn{1}{c|}{0.546}          & \multicolumn{1}{c|}{0.269}          & \multicolumn{1}{c|}{0.856}          & 0.622          & \textbf{0.925}            \\ \hline
Geonrw                  & MCC Transfer Learning                     & \multicolumn{1}{c|}{0.624}          & \multicolumn{1}{c|}{0.584}          & \multicolumn{1}{c|}{0.406}          & \multicolumn{1}{c|}{0.847}          & \textbf{0.659} & 0.877                     \\ \hline
DeepGlobe               & Combined Training                  & \multicolumn{1}{c|}{0.612}          & \multicolumn{1}{c|}{0.601}          & \multicolumn{1}{c|}{0.409}          & \multicolumn{1}{c|}{0.822}          & 0.616          & 0.901                     \\ \hline
DeepGlobe               & MCC Semi-Supervised           & \multicolumn{1}{c|}{0.615}          & \multicolumn{1}{c|}{0.615}          & \multicolumn{1}{c|}{0.368}          & \multicolumn{1}{c|}{0.823}          & \textbf{0.653} & 0.887                     \\ \hline
DeepGlobe               & MCC Transfer Learning                     & \multicolumn{1}{c|}{0.533}          & \multicolumn{1}{c|}{0.583}          & \multicolumn{1}{c|}{0.121}          & \multicolumn{1}{c|}{0.837}          & 0.593          & \textbf{0.923}            \\ \hline
\end{tabular}
}
\label{tab:tab2}
\end{center}
\end{table*}

The first row in the table \ref{tab:tab2} gives the results obtained from the supervised learning problem without using any source data. HWLC18 database, which is in the source dataset, is labeled similarly to the HWLC16. The biggest difference between the two datasets is the city/rural area differences they cover and the size of the images.  As seen in experiments with HWLC18 dataset without resizing, semi-supervised learning provides the greatest performance increase using only images without labels. Combined training and transfer learning cannot increase prediction performance due to differences in the distribution of labels, reducing it from 60.2\% to 59.3\% in rural areas and from 79.6\% to 51.8\% in urban areas. \\

We see that the transfer learning performance exceeds both methods in the experiments where we equate the image and label sizes, which is the main difference between these two datasets, and show them in the table with HWLC18-$>$16. In these experiments, the success reached the highest at 63.6\%, especially in rural areas targeted by the HWLC16 dataset. \\ 

Transfer learning experiments on the Geonrw dataset showed us the effect of differences between labelings on transfer success from different performances in rural and urban regions. While the Geonrw database is labeled with similar annotation rules as HWLC16 in land cover classes, it shows significant differences in the labeling of building regions. The areas between buildings are labeled as buildings due to low resolution in HWLC16. Geonrw has more detailed annotations in areas such as industrial regions and airports. The differences between the class definitions caused transfer to perform worse than semi-supervised learning, especially in the urban dataset, while in rural areas, semi-supervised learning performed about 5\% below the combined training and transfer learning. \\

Similarly, experiments with DeepGlobe dataset resulted in a 92.3\% transfer learning success because the building labels in this dataset are similar to our problem. However, we suspect that the differences between the target areas and labelings of the datasets cause the targeted performance in rural areas to be lower than other datasets.\\

We experienced HRNet model to achieve building segmentation performance comparison with other models. This model achieved the best performance in city regions, not in rural area buildings. As you see in Table \ref{tab:tab3} HRNet achieved nearly 10 \% increment in IoU score in building segmentation. This model seems promising to us, and we will dive into it for our next studies. 
\begin{table}[htbp]
\caption{The building segmentation results of the experiments performed on the HWLC16 dataset with the architectures of different models.}
\begin{center}
\resizebox{.99\hsize}{!}{$
\begin{tabular}{| c | c | c |  c |}
\hline
\textbf{Model} & \textbf{Encoder} & \textbf{Rural Area Dataset}	& \textbf{City} \\ \hline
               		&                    		 & \textbf{Village}        	& \textbf{Building}      \\ \hline
BisenetV1      	& Resnet18           	 &0.55           	          	& 0.721                     \\ \hline
Unet++          	& Resnet50           	 & \textbf{0.613}               & 0.76                      \\ \hline
Unet++          	& EfficientnetB3    	 & 0.595          	 	 	& 0.796            		\\ \hline
Unet++          	& EfficientnetB5    	 & 0.582         	         	& 0.787                     \\ \hline
Deeplab+       	& Resnet50          	 & 0.471          	         	& 0.703                     \\ \hline
Deeplab+       	& EfficientnetB3       	 & 0.584          	         	& 0.68                      \\ \hline
Deeplab+       	& EfficientnetB5   	 & 0.582          	        	& 0.762                     \\ \hline
HRNet         	& hrnetv2w40    	 & 0.578          	         	& \textbf{0.863}          \\ \hline
\end{tabular}\label{tab:tab3}$}
\end{center}
\end{table}

\section{Conclusions}

In this study, we present the results of our research on the use of semantic segmentation methods with various multi-source prediction methods for extracting of land cover maps from satellite images on the HWLC16 and HWLC18 datasets. \\

In the experiments, the segmentation success of urban areas 79.6\%  and rural areas 60.2\%  was reached with the original resolution in the HWLC16 dataset. The addition of the high-resolution HWLC18 dataset, which has been taken from various sources and has the same annotation rules, through transfer learning increased segmentation performance to 90.1 \% for urban areas and  63.6\% for rural areas in the same test set. Geonrw and DeepGlobe databases are used in training using a semi-supervised learning method due to the fact that although these datasets have high resolution, are annotated with different labeling rules and classes. The possible reason for this can be transfer between these tasks leads to negative transfer. For these experiments, the performance in the urban area dataset increased up to 92.3\%. \\

In future work, we aim to make improvements to the diversification of transfer learning and semi-supervised learning methods and to improve the minimum class confusion method designed for classification by subjecting it to appropriate normalizations in accordance with the segmentation problem. Indeed we will modify minimum class confusion loss to be able to change priority according to epochs. Additional experiments with HRNet showed us that multi-scale fusion between encoder and decoder architectures in an end-to-end manner improved semantic segmentation results dramatically. To conclude, we will conduct more experiments with this architecture to improve semantic segmentation results on both rural and urban areas by using several transfer learning approaches.


\nocite{*} 
\begin{thebibliography}{10}
\providecommand{\url}[1]{#1}
\csname url@samestyle\endcsname
\providecommand{\newblock}{\relax}
\providecommand{\bibinfo}[2]{#2}
\providecommand{\BIBentrySTDinterwordspacing}{\spaceskip=0pt\relax}
\providecommand{\BIBentryALTinterwordstretchfactor}{4}
\providecommand{\BIBentryALTinterwordspacing}{\spaceskip=\fontdimen2\font plus
\BIBentryALTinterwordstretchfactor\fontdimen3\font minus
  \fontdimen4\font\relax}
\providecommand{\BIBforeignlanguage}[2]{{%
\expandafter\ifx\csname l@#1\endcsname\relax
\typeout{** WARNING: IEEEtran.bst: No hyphenation pattern has been}%
\typeout{** loaded for the language `#1'. Using the pattern for}%
\typeout{** the default language instead.}%
\else
\language=\csname l@#1\endcsname
\fi
#2}}
\providecommand{\BIBdecl}{\relax}
\BIBdecl

\bibitem{hrnet}
\BIBentryALTinterwordspacing
J.~Wang, K.~Sun, T.~Cheng, B.~Jiang, C.~Deng, Y.~Zhao, D.~Liu, Y.~Mu, M.~Tan,
  X.~Wang, W.~Liu, and B.~Xiao, ``Deep high-resolution representation learning
  for visual recognition,'' \emph{CoRR}, vol. abs/1908.07919, 2019. [Online].
  Available: \url{http://arxiv.org/abs/1908.07919}
\BIBentrySTDinterwordspacing

\bibitem{druck-etal-2009-active}
G.~Druck, B.~Settles, and A.~McCallum, ``Active learning by labeling
  features,'' in \emph{Conference on Empirical Methods in Natural Language
  Processing}, Aug. 2009, pp. 81--90.

\bibitem{long2015fully}
J.~Long, E.~Shelhamer, and T.~Darrell, ``Fully convolutional networks for
  semantic segmentation,'' in \emph{Proceedings of the IEEE conference on
  computer vision and pattern recognition}, 2015, pp. 3431--3440.

\bibitem{ronneberger2015u}
O.~Ronneberger, P.~Fischer, and T.~Brox, ``U-net: Convolutional networks for
  biomedical image segmentation,'' in \emph{MICCAI2015}.\hskip 1em plus 0.5em
  minus 0.4em\relax Springer, 2015.

\bibitem{zhou2018unet++}
Z.~Zhou, M.~M. Rahman~Siddiquee, N.~Tajbakhsh, and J.~Liang, ``Unet++: A nested
  u-net architecture for medical image segmentation,'' in \emph{Deep learning
  in medical image analysis and multimodal learning for clinical decision
  support}.\hskip 1em plus 0.5em minus 0.4em\relax Springer, 2018, pp. 3--11.

\bibitem{chen2014semantic}
L.-C. Chen, G.~Papandreou, I.~Kokkinos, K.~Murphy, and A.~L. Yuille, ``Semantic
  image segmentation with deep convolutional nets and fully connected crfs,''
  \emph{arXiv preprint arXiv:1412.7062}, 2014.

\bibitem{chen2017deeplab}
L.-C. Chen, G.~Papandreou, I.~Kokkinos, and K.~Murphy, ``Deeplab: Semantic
  image segmentation with deep convolutional nets, atrous convolution, and
  fully connected crfs,'' \emph{IEEE transactions on PAMI}, vol.~40, no.~4, pp.
  834--848, 2017.

\bibitem{chen2017rethinking}
L.-C. Chen, G.~Papandreou, F.~Schroff, and H.~Adam, ``Rethinking atrous
  convolution for semantic image segmentation,'' \emph{arXiv preprint
  arXiv:1706.05587}, 2017.

\bibitem{chen2018encoder}
L.-C. Chen, Y.~Zhu, G.~Papandreou, F.~Schroff, and H.~Adam, ``Encoder-decoder
  with atrous separable convolution for semantic image segmentation,'' in
  \emph{ECCV}, 2018, pp. 801--818.

\bibitem{neupane2021deep}
B.~Neupane, T.~Horanont, and J.~Aryal, ``Deep learning-based semantic
  segmentation of urban features in satellite images: A review and
  meta-analysis,'' \emph{Remote Sensing}, vol.~13, no.~4, p. 808, 2021.

\bibitem{guo2018semantic}
Z.~Guo, H.~Shengoku, G.~Wu, Q.~Chen, W.~Yuan, X.~Shi, X.~Shao, Y.~Xu, and
  R.~Shibasaki, ``Semantic segmentation for urban planning maps based on
  u-net,'' in \emph{IGARSS 2018-2018 IEEE International Geoscience and Remote
  Sensing Symposium}.\hskip 1em plus 0.5em minus 0.4em\relax IEEE, 2018, pp.
  6187--6190.

\bibitem{li2019semantic}
W.~Li, C.~He, J.~Fang, J.~Zheng, H.~Fu, and L.~Yu, ``Semantic
  segmentation-based building footprint extraction using very high-resolution
  satellite images and multi-source gis data,'' \emph{Remote Sensing}, vol.~11,
  no.~4, p. 403, 2019.

\bibitem{yi2019semantic}
Y.~Yi, Z.~Zhang, W.~Zhang, and C.~Zhang, ``Semantic segmentation of urban
  buildings from vhr remote sensing imagery using a deep convolutional neural
  network,'' \emph{Remote sensing}, vol.~11, no.~15, 2019.

\bibitem{du2021incorporating}
S.~Du, S.~Du, B.~Liu, and X.~Zhang, ``Incorporating deeplabv3+ and object-based
  image analysis for semantic segmentation of very high resolution remote
  sensing images,'' \emph{International Journal of Digital Earth}, vol.~14,
  no.~3, pp. 357--378, 2021.

\bibitem{hong2021spectralformer}
D.~Hong, Z.~Han, J.~Yao, L.~Gao, B.~Zhang, A.~Plaza, and J.~Chanussot,
  ``Spectralformer: Rethinking hyperspectral image classification with
  transformers,'' \emph{IEEE Transactions on Geoscience and Remote Sensing},
  vol.~60, pp. 1--15, 2021.

\bibitem{xu2021efficient}
Z.~Xu, W.~Zhang, Z.~Yang, and J.~Li, ``Efficient transformer for remote sensing
  image segmentation,'' \emph{Remote Sensing}, vol.~13, p. 3585, 2021.

\bibitem{scheibenreif2022self}
L.~Scheibenreif, J.~Hanna, M.~Mommert, and D.~Borth, ``Self-supervised vision
  transformers for land-cover segmentation and classification,'' in
  \emph{CVPR}, 2022, pp. 1422--1431.

\bibitem{baier2020building}
G.~Baier, A.~Deschemps, M.~Schmitt, and N.~Yokoya, ``Building a parallel
  universe—image synthesis from land cover maps and auxiliary raster data,''
  \emph{CoRR}, 2020.

\bibitem{maggiori2017dataset}
E.~Maggiori, Y.~Tarabalka, G.~Charpiat, and P.~Alliez, ``The inria aerial image
  labeling benchmark,'' in \emph{IEEE International Geoscience and Remote
  Sensing Symposium (IGARSS)}.\hskip 1em plus 0.5em minus 0.4em\relax IEEE,
  2017.

\bibitem{demir2018deepglobe}
I.~Demir, K.~Koperski, D.~Lindenbaum, G.~Pang, J.~Huang, S.~Basu, F.~Hughes,
  D.~Tuia, and R.~Raskar, ``Deepglobe 2018: A challenge to parse the earth
  through satellite images,'' in \emph{Proceedings of the IEEE CVPR Workshops},
  2018, pp. 172--181.

\bibitem{wang2018deep}
M.~Wang and W.~Deng, ``Deep visual domain adaptation: A survey,''
  \emph{Neurocomputing}, vol. 312, pp. 135--153, 2018.

\bibitem{Zhang2019}
J.~Zhang, W.~Li, P.~Ogunbona, and D.~Xu, ``Recent advances in transfer learning
  for cross-dataset visual recognition: A problem-oriented perspective,''
  \emph{ACM Computing Surveys}, vol.~52, pp. 1--38, 2019.

\bibitem{ouali2020overview}
Y.~Ouali, C.~Hudelot, and M.~Tami, ``An overview of deep semi-supervised
  learning,'' \emph{arXiv preprint arXiv:2006.05278}, 2020.

\bibitem{wilson2020survey}
G.~Wilson and D.~J. Cook, ``A survey of unsupervised deep domain adaptation,''
  \emph{ACM Transactions on Intelligent Systems and Technology (TIST)},
  vol.~11, no.~5, pp. 1--46, 2020.

\bibitem{hinton2007backpropagation}
G.~Hinton \emph{et~al.}, ``How to do backpropagation in a brain,'' in
  \emph{Invited talk at the NIPS’2007 deep learning workshop}, vol. 656,
  2007, pp. 1--16.

\bibitem{jin2020minimum}
Y.~Jin, X.~Wang, M.~Long, and J.~Wang, ``Minimum class confusion for versatile
  domain adaptation,'' in \emph{European Conference on Computer Vision}.\hskip
  1em plus 0.5em minus 0.4em\relax Springer, 2020, pp. 464--480.

\bibitem{long2017deep}
M.~Long, H.~Zhu, J.~Wang, and M.~I. Jordan, ``Deep transfer learning with joint
  adaptation networks,'' in \emph{ICML}, 2017.

\bibitem{li2016revisiting}
Y.~Li, N.~Wang, J.~Shi, J.~Liu, and X.~Hou, ``Revisiting batch normalization
  for practical domain adaptation,'' \emph{arXiv:1603.04779}, 2016.

\bibitem{cubuk2020randaugment}
E.~D. Cubuk, B.~Zoph, J.~Shlens, and Q.~V. Le, ``Randaugment: Practical
  automated data augmentation with a reduced search space,'' in \emph{CVPRW},
  2020, pp. 702--703.

\bibitem{xie2020self}
Q.~Xie, M.-T. Luong, E.~Hovy, and Q.~V. Le, ``Self-training with noisy student
  improves imagenet classification,'' in \emph{CVPR}, 2020.

\bibitem{tan2019efficientnet}
M.~Tan and Q.~Le, ``Efficientnet: Rethinking model scaling for convolutional
  neural networks,'' in \emph{ICML}.\hskip 1em plus 0.5em minus 0.4em\relax
  PMLR, 2019, pp. 6105--6114.

\bibitem{Openstreetmaps:2022}
\BIBentryALTinterwordspacing
{Openstreetmap contributors}, ``Zoom levels \- openstreetmap,'' 2022, [Online;
  accessed 6-June-2022]. [Online]. Available:
  \url{https://wiki.openstreetmap.org/wiki/Zoom\_levels}
\BIBentrySTDinterwordspacing

\bibitem{geonrw-20}
G.~Baier, A.~Deschemps, M.~Schmitt, and N.~Yokoya, ``Geonrw,'' 2020.

\bibitem{Yakubovskiy:2019}
\BIBentryALTinterwordspacing
P.~Yakubovskiy, ``Segmentation models pytorch,'' 2020. [Online]. Available:
  \url{https://github.com/qubvel/segmentation\_models.pytorch}
\BIBentrySTDinterwordspacing

\end{thebibliography}


\end{document}